%% file: iclr2027_conference.tex
\documentclass{article} %
\usepackage[T1]{fontenc}
\PassOptionsToPackage{dvipsnames}{xcolor} %
\usepackage{iclr2027_conference,times}

\input{math_commands.tex}

\usepackage{amsmath,amssymb}
\usepackage{booktabs}
\usepackage{multirow}
\usepackage{makecell}
\usepackage{graphicx}
\usepackage{wrapfig}
\usepackage{needspace}
\usepackage{tikz}
\usetikzlibrary{arrows.meta,positioning,calc}
\usepackage{colortbl} %
\usepackage{algorithm}
\usepackage{algorithmic}
\usepackage{hyperref}
\usepackage{url}
\usepackage{pifont}
\usepackage{placeins}
\usepackage{float}
\usepackage{enumitem}
\usepackage{tcolorbox}

\newcommand{\method}{\textsc{ReCAP}}

\title{Persistent Context Graphs for \\ Efficient Memory Compaction in LLM Agents}

\author{Jingbo Yang$^{1}$\thanks{Correspondence to: Jingbo Yang\texttt{<jingbo@ucsb.edu>}.}\quad Kwei-Herng Lai$^{2}$\quad Xiaowen Wang$^{2}$\quad Zhaoxuan Tan$^{3}$\quad Pei Zhou$^{2}$\\
\textbf{Mengting Wan$^{2}$\quad Yaar Harari$^{2}$\quad Evgeniy Gabrilovich$^{2}$\quad Shiyu Chang$^{1}$} \\
$^{1}$University of California, Santa Barbara \quad
$^{2}$Microsoft \quad
$^{3}$University of Notre Dame \\
}

\iclrfinalcopy %
\begin{document}

\maketitle

\begin{abstract}
As LLM capabilities advance, agents are tackling increasingly complex tasks over longer horizons. Their growing interaction histories make memory compaction essential for staying within context windows and reducing prefill cost. Existing methods summarize the history or compress its KV cache, often adding model computation to preserve information for future requests. A new user request can change which history matters, but reassessing that history with the model requires re-encoding it if the KV cache has expired.
Past attention provides signals of historical importance and dependencies between messages, while relevance to the current task must be assessed using the new user request. We introduce \method{}, a memory compaction method that stores attention-derived importance scores and dependency links in a lightweight, persistent context graph. For each new request, \method{} combines stored importance with relevance cues from the request and follows dependency links to select messages and their supporting context, without additional model calls for selection.
Compared with Codex's default summarization-based compaction, \method{} reduces estimated latency for compaction and cold restoration by approximately $95\%$ on both Qwen3-Coder and gpt-oss. It also roughly halves the historical context per call on SWE-Together at comparable task quality and improves accuracy on the code tasks of Lost-in-Conversation over full history by $19.8$ and $41.2$ points. Code is available at \url{https://github.com/UCSB-NLP-Chang/ReCAP}.
\end{abstract}

\section{Introduction}
\label{sec:intro}

Advances in LLM capabilities are enabling agents to solve increasingly complex tasks over longer horizons (\emph{e.g.}, repository-level coding and autonomous research). Such workflows can run for tens of minutes to hours, accumulating user instructions, intermediate decisions, and tool outputs as execution proceeds. The resulting history can eventually exceed the model's finite context window. It also raises inference cost: measurements of agentic workloads report inputs tens of times longer than their generated outputs, making repeated context encoding a substantial part of execution~\citep{lu2026mix}. Modern agent harnesses and recent research therefore adopt \emph{memory compaction}, which consolidates session memory into a shorter context~\citep{kang2025acon}. By limiting the history carried into subsequent calls, compaction allows the agent to continue beyond a single context window while reducing repeated prefilling.

Existing approaches compact agent memory either by summarizing the history into a shorter prompt or by retaining a subset or compressed form of the model's KV cache~\citep{kang2025acon,li2024snapkv,zweiger2026fast}. Doing so efficiently across an interactive session involves three linked challenges.
\ding{182}~\textbf{What to retain depends on future use.} A new user request can change the task's direction or revisit an earlier decision, so compaction performed before it arrives must preserve information for needs not yet known. Query-aware KV eviction, for instance, degrades when later queries differ from the one that guided eviction~\citep{li2025scbench,kim2026kvzip}.
\ding{183}~\textbf{Anticipating future use requires additional computation.} Deciding which history a later request will use requires queries that resemble that request, which do not yet exist when compaction runs. Attention Matching~\citep{zweiger2026fast}, for example, synthesizes such reference queries by re-prefilling the context or simulating interactions and fits the compact cache to them, adding model computation to every compaction.
\ding{184}~\textbf{Cache expiration adds reconstruction cost.} In about $4{,}300$ Claude Code and Codex sessions, user think time accounts for $92\%$ of session time, and requests after hour-long gaps almost always miss the serving cache~\citep{zhu2026tracelab}. Compacted KV states must then be stored and reloaded~\citep{gao2025fast}, and a compactor that waits for the request must re-encode the history to obtain fresh attention.

\begin{figure}[!t]
\centering
\includegraphics[width=\textwidth]{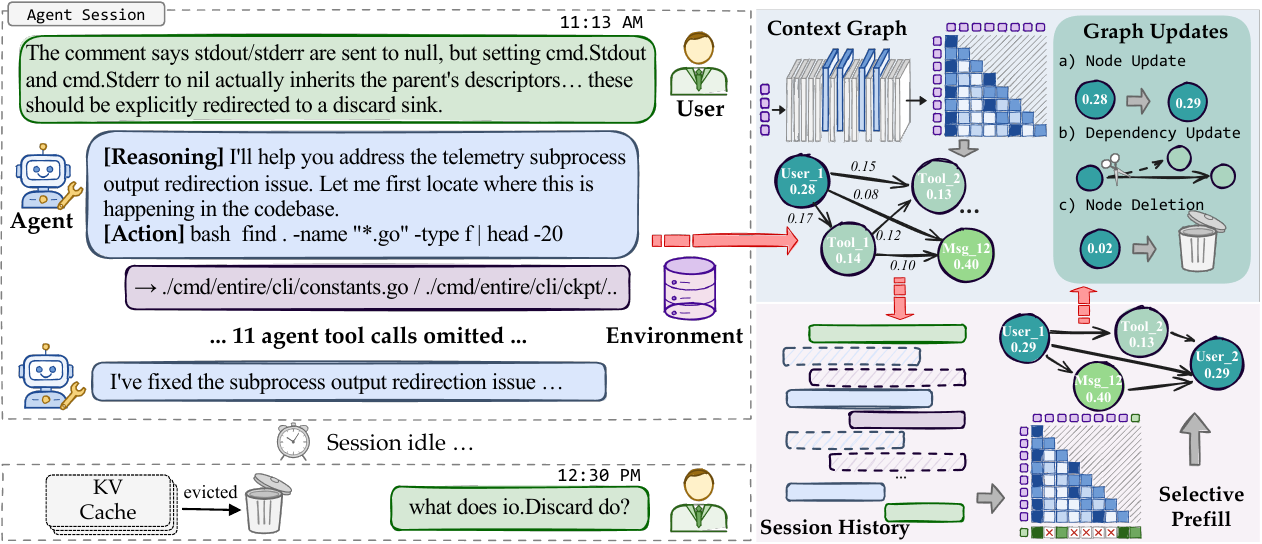}
\caption{\method{} preserves attention-derived importance and dependency links in a persistent context graph. For each new request, it combines stored importance with the request's relevance cues to select historical blocks and their supporting context, retaining protected contexts.}
\label{fig:overview}
\end{figure}

Each forward pass of the agent already computes attention over its history, yet this signal is discarded once the request completes. We study how coding agents attend to their histories across user turns and find that attention concentrates on a small part of the history, that past attention predicts which history the next turn reuses, and that the arriving request redirects attention (\S\ref{sec:observations}). In light of these findings, we propose \method{}, a memory compaction method that keeps this attention rather than discarding it and organizes it in a lightweight, \emph{persistent context graph} (Figure~\ref{fig:overview}): nodes are history blocks annotated with accumulated importance, and edges link each block to the earlier blocks it depends on. When a new request arrives, \method{} selects from the full history by combining stored importance with the request's identifier overlap and following dependency edges, so selection adapts to the actual request and can restore previously omitted blocks~(\ding{182}). The graph is built from attention that inference computes anyway, and selection uses only graph traversal and string matching, so compaction adds no model call~(\ding{183}). Because the graph persists independently of the KV cache, \method{} can select after the cache expires without re-encoding the full history~(\ding{184}).

Our contributions are as follows:
\begin{itemize}[leftmargin=*,itemsep=1pt,topsep=2pt,parsep=0pt]
\item We show that attention from ordinary agent inference concentrates on a small part of the history, predicts which history the next turn reuses, and is redirected by the arriving request (\S\ref{sec:observations}).
\item We propose \method{}, which persists this attention in a context graph of block importance and dependencies and queries it for each new request to select context without an additional model call, independently of the KV cache (\S\ref{sec:method}).
\item On SWE-Together~\citep{wu2026swe} and the code tasks of Lost-in-Conversation~\citep{laban2026llms} with two model families, \method{} reduces the estimated latency of compaction and cold restoration by approximately $95\%$ relative to Codex's default summarization, roughly halves the historical context on SWE-Together at comparable quality, and improves Lost-in-Conversation accuracy over full-history conditioning by $19.8$ and $41.2$ points (\S\ref{sec:experiments}).
\end{itemize}

\section{Related Work}
\label{sec:related}

\textbf{Context management for LLM agents.}
Agent harnesses summarize long histories with the backbone, and recent work optimizes such compressors or trains agents to fold their own context~\citep{kang2025acon,sun2025scaling,zhou2026mem1}; both add model computation and discard detail irreversibly. Cheaper alternatives mask old observations~\citep{lindenbauer2025complexity} or prune text with small models~\citep{pan2024llmlingua,jiang2024longllmlingua}. Two recent methods select history by dependencies or attention at higher cost. ContextWeaver calls the backbone at every step to identify the earlier steps it depends on and to summarize these dependencies, and it replaces the observations of all other steps with placeholders~\citep{wu2026contextweaver}. AttnCompress reruns a separate 4B-parameter model over the trajectory and scores blocks by the attention of the agent's next generation step, recomputing all scores at each refresh~\citep{zeng2026attncompress}. Both target autonomous single-issue runs, in which no new user request redirects the task. \method{} records importance and dependencies from the agent's own attention during inference and accumulates them across turns. When a user request arrives, it selects from the full archive without a model call, so omitted blocks can return.

\textbf{KV-cache compression and reuse.}
KV methods evict cached entries by attention statistics~\citep{zhang2023h2o,li2024snapkv,feng2026ada} or fit compact states~\citep{zweiger2026fast}, and query-aware eviction degrades when later queries differ~\citep{li2025scbench,kim2026kvzip}. In agents, compacting each turn immediately loses accuracy, while waiting for later queries requires keeping the uncompressed cache meanwhile~\citep{liu2026practical}; learned cross-turn scoring still evicts entries permanently~\citep{li2026intentkv}. Recallable methods and serving systems avoid such loss by keeping full caches or session state in GPU or host memory~\citep{tang2024quest,xiao2024infllm,gao2024cost,gao2025fast}. Yet user think time dominates coding-agent sessions, and caches often expire across these gaps~\citep{zhu2026tracelab}. \method{} waits for the actual request without holding KV state: its text archive and graph survive cache expiration, and it coexists with prefix caching.

\section{Memory Compaction: Setting and Observations}
\label{sec:problem}

\subsection{Query-conditioned memory compaction}
\label{sec:setting}

An agent session alternates between user requests and agent execution. In turn $t$, the user sends a request $q_t$, and the agent produces assistant messages, tool calls, and tool results before returning control; $H_t$ denotes the accumulated history. When $q_{t+1}$ arrives, memory compaction constructs a smaller working context from $H_t$ for the new request, limiting the history carried into subsequent calls and thus both context-window pressure and encoding cost. If the previous KV state has expired or been evicted, the engine rebuilds it by prefilling this compact context.

We study compaction by selecting historical message blocks. Let $V_t$ contain the selectable blocks in $H_t$, excluding system instructions (\S\ref{sec:segmentation}), and let $C\subseteq V_t$ be the selected set, with cost $c(C)=\sum_{v\in C}c(v)$ in characters of message content and serialized tool calls. A nominal retention fraction $\beta\in(0,1]$ sets the packing budget $B_t=\lfloor\beta\,c(V_t)\rfloor$. System instructions and the current turn are rendered separately. Our method prioritizes protected blocks over this nominal budget and allows bounded dependency expansion (\S\ref{sec:selection}); we measure the resulting token count, latency, and quality.

We seek a compactor that \emph{(i)}~selects context from stored information without an additional model call when a request arrives, \emph{(ii)}~keeps tool calls paired with their results, \emph{(iii)}~keeps the selected historical prefix fixed across tool steps within a user turn for prefix-cache reuse, and \emph{(iv)}~adapts to the arriving request, including requests that revisit earlier work.
These requirements separate two sources of evidence: historical usage becomes available during execution, whereas relevance to the next request can be assessed only after that request arrives. A reusable representation of past usage can connect these two stages, even when the KV state is no longer available.

\subsection{What attention reveals about useful context}
\label{sec:measure}
\label{sec:observations}

We examine whether attention provides a historical importance signal, whether that signal predicts later reuse, and how the arriving request changes access to the same history, using a multi-turn example, a cross-turn reuse study, and a controlled query study (Figure~\ref{fig:obs}; protocols in Appendix~\ref{app:measure}).

\begin{figure}[t]
\centering
\includegraphics[width=\textwidth]{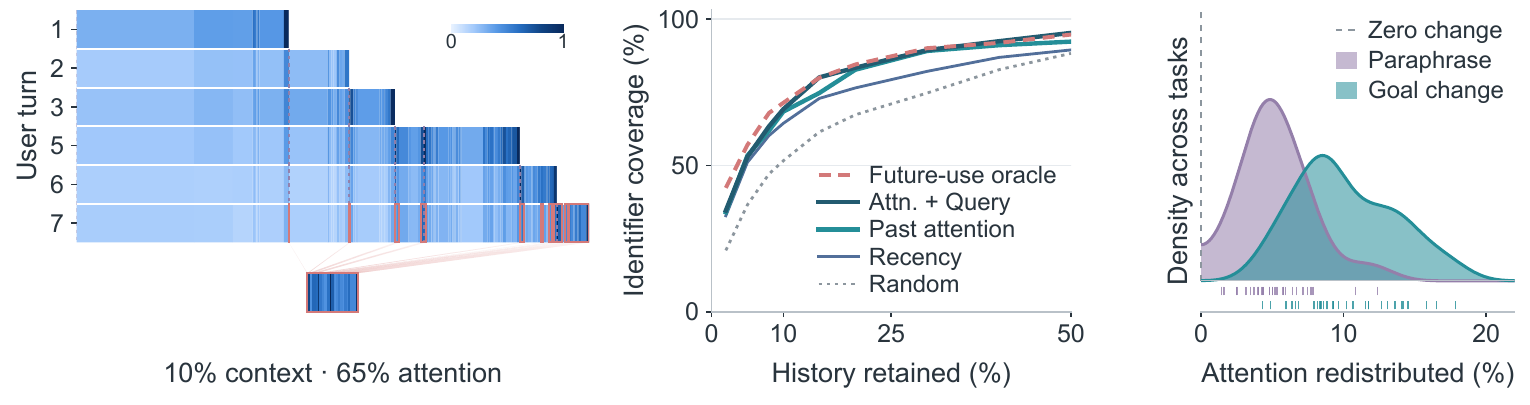}
\vspace{-15pt}
\caption{\textbf{Attention signals for memory compaction.}
\textbf{Left:} Attention over history blocks per turn (width: tokens; dashed: user turns; blue: attention density, cube-root scale). Coral: blocks selected under a $10\%$ token budget, concatenated below.
\textbf{Middle:} Next-turn identifier coverage versus retained-history budget; \emph{Attn.\ + Query} adds query-identifier overlap.
\textbf{Right:} Attention redistribution (TV distance) over $32$ fixed histories for paraphrases versus changed targets; ticks mark tasks.}
\label{fig:obs}
\end{figure}

\textbf{Observation 1: attention concentrates on a small part of the history.}
In one coding session (Figure~\ref{fig:obs}, left), attention is unevenly distributed over historical blocks at every checkpoint: ranking blocks by attention per token within a $10\%$ historical-token budget retains $65\%$ of the attention mass at the final checkpoint. The reuse-study transitions show the same concentration (Appendix~\ref{app:measure}). This concentration gives a compactor a ranking signal for allocating a limited context budget.

\textbf{Observation 2: past attention predicts subsequent reuse.}
We evaluate selection signals over $64$ turn transitions from eight coding-agent sessions, approximating reuse by the IDF-weighted coverage of identifiers that appear during the next turn's execution. At a $10\%$ historical-text budget, ranking by attention from the preceding turn reaches a median coverage of about $0.68$, close to the $0.71$ of a future-use oracle that ranks blocks by those identifiers (Figure~\ref{fig:obs}, middle). Past attention thus supplies an importance signal that remains useful beyond the turn in which it was measured. Adding overlap with the incoming query's identifiers yields about $0.69$. We next hold the history fixed to isolate the request's effect on attention.

\textbf{Observation 3: the arriving request redirects attention.}
For each of $32$ tasks, we hold one historical context fixed and vary only the request, using two phrasings for each of two target files, and measure attention change over historical blocks by total-variation distance. Changing the target redistributes $10.1\%$ of attention on average, compared with $5.3\%$ for paraphrasing (Figure~\ref{fig:obs}, right), and produces greater redistribution in $28$ of the $32$ tasks. This pattern also holds for requests that explicitly ask the agent to recall earlier work (Appendix~\ref{app:query-study}). Historical importance therefore supports selection across turns, while the arriving request adds a cue for relevance to the current task.

These observations motivate preserving historical importance between turns and combining it with the current request when selecting context. \method{} does so with a persistent context graph: message nodes retain attention-derived importance, and attention links provide dependency cues for supporting context, independently of the KV cache.

\section{\method{}: Memory Compaction with a Persistent Context Graph}
\label{sec:method}

\method{} maintains a persistent context graph that connects past attention to future compaction decisions: node scores summarize historical importance, and directed edges record attention-based dependencies between messages. It \emph{updates} the graph using attention from agent execution and \emph{queries} it when a new user request arrives, combining stored importance with relevance to that request and including supporting context through graph traversal. Figure~\ref{fig:method} illustrates the workflow; Algorithm~\ref{alg:recap} in Appendix~\ref{app:graph-algorithm} gives the complete procedure.

\begin{figure}[!t]
\centering
\includegraphics[width=\textwidth]{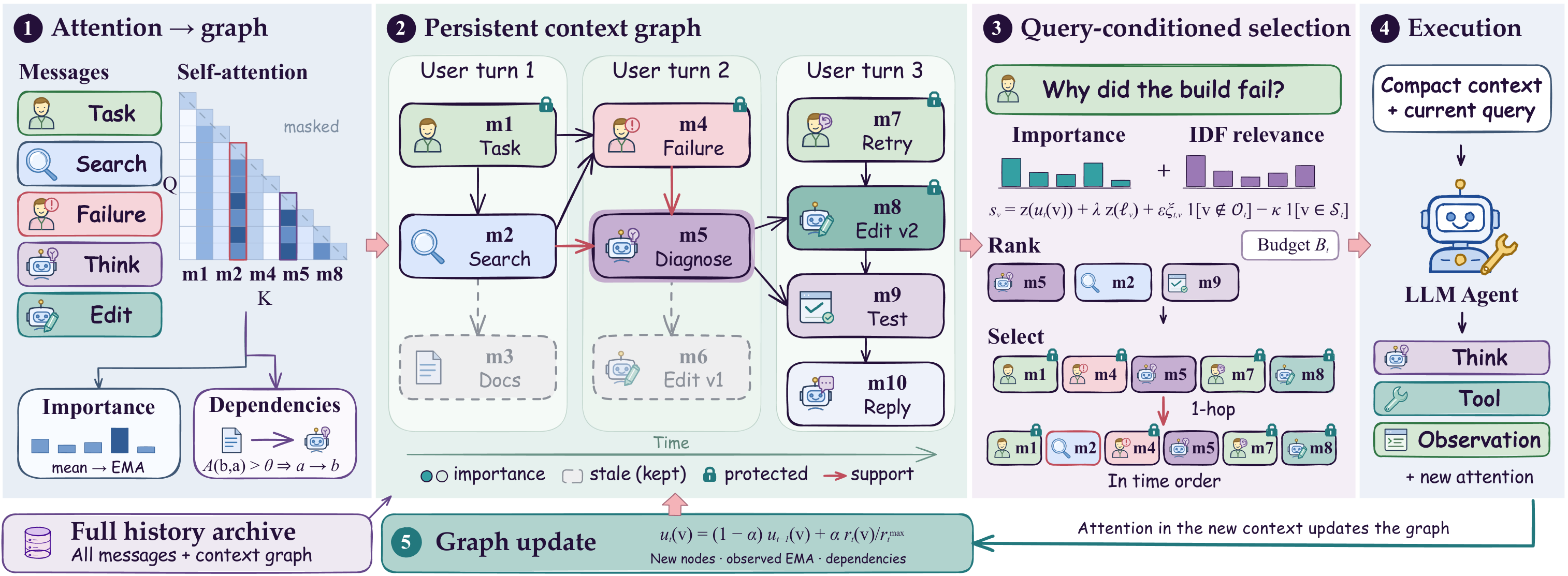}
\vspace{-15pt}
\caption{\textbf{ReCAP workflow.} \textbf{1--2:} Sampled attention supplies node importance and support-to-dependent edges. \textbf{3:} Selection combines stored importance with the query's identifier overlap, keeps protected records, and adds one-hop predecessors (diagnosis $m_5$ brings in search record $m_2$), without a model call. \textbf{4--5:} Execution produces new messages and attention that update the graph.}
\label{fig:method}
\end{figure}

\subsection{Modeling Session History as an Evolving Context Graph}
\label{sec:segmentation}

Each node represents an \emph{atomic block} of history that can be carried into the working context: a user message, an assistant message that issues tool calls together with their results, or a plain assistant message. Grouping tool calls with their results preserves a valid message sequence when blocks are selected. Nodes receive stable identifiers in chronological order as the history grows.

We define the context graph as a directed graph $G_t=(V_t,E_t)$ that evolves with each turn, where each node $v\in V_t$ is an atomic block and carries an importance score $u_t(v)\in[0,1]$ as a node attribute. Each edge $e=(a,b)\in E_t$ is an ordered pair from an earlier supporting block $a$ to a later dependent block $b$ and is associated with a weight $w_t(a,b)>0$ that indicates the strength of the dependency: selecting a diagnosis, for example, can bring in the failure trace it relies on. The graph stores block references, scalar scores, and dependency links alongside the full history. These records persist across turns independently of the KV cache, so later requests can select any accumulated block, even one previously omitted.

\subsection{Persisting importance and dependencies from attention}
\label{sec:feedback}

At each turn boundary, \method{} consolidates attention over the working context to node scores and dependency links. Let $R_t\subseteq V_t$ be the blocks observed in this update, including those created during the turn. Attention from the completed turn measures each block's use, and attention between blocks identifies supporting relationships; both use sampled attention rows (Appendix~\ref{app:implementation}).

\textbf{Accumulating historical importance.}
We sample token positions $W_t$ from the completed turn and average their attention over heads and selected layers; after clipping extreme values, $\rho_t(j)$ denotes the attention received by historical token $j$. For each block $v\in R_t$, we average $\rho_t$ over its tokens $\mathrm{span}_t(v)$ and update its importance with an exponential moving average:
\begin{equation}
\bar\rho_t(v)=\frac{1}{|\mathrm{span}_t(v)|}\sum_{j\in\mathrm{span}_t(v)}\rho_t(j),
\qquad
u_t(v)=(1-\alpha)u_{t-1}(v)+\alpha\frac{\bar\rho_t(v)}{\bar\rho_t^{\max}},
\label{eq:utility}
\end{equation}
where $\bar\rho_t^{\max}=\max_{v\in R_t}\bar\rho_t(v)$ and new nodes start at zero. Per-token averaging accounts for block length, while the moving average combines recent usage with previous observations. Nodes outside $R_t$ retain their stored scores, keeping this evidence for later requests that revisit omitted history.

\textbf{Recording dependencies.}
To identify the context supporting a block $b$, we sample positions within $b$ and measure the mean attention mass they assign to each earlier block $a$ at a selected layer. When this mass exceeds a threshold $\theta$, we add the edge $(a,b)$, pointing from support to dependent (the reverse of the attention direction). Edges accumulate across turns, and each weight $w_t(a,b)$ keeps the largest mass observed so far. During compaction, they let a selected block retrieve its supporting predecessors, including relationships formed within a single turn.

The graph also tracks observation status and potential supersession. The set $\mathcal{O}_t=\mathcal{O}_{t-1}\cup R_t$ records which nodes have received attention measurements, and a stale set $\mathcal{S}_t$ marks earlier blocks whose identifiers substantially overlap a later block that receives stronger attention (Appendix~\ref{app:graph-algorithm}). These marks lower the priority of potentially superseded content while preserving its node and text.

\subsection{Querying the graph for memory compaction}
\label{sec:selection}

When $q_{t+1}$ arrives, \method{} queries $G_t$ to construct a compact context for it. Historical importance supplies the prior usage signal established in \S\ref{sec:observations}, and the request supplies the relevance cue. Selection combines these signals to rank nodes, packs them with protected records, and follows dependency edges to add supporting context, using only stored graph state and string matching.

\textbf{Query-conditioned node scores.}
We compute a relevance cue from identifier overlap between the request and each block, weighted by inverse document frequency:
\begin{equation}
\ell_v=\sum_{w\in I(q_{t+1})\cap I(v)}\log\frac{|V_t|}{\mathrm{df}_t(w)},
\label{eq:lexical}
\end{equation}
where $I(\cdot)$ extracts identifiers from message text and $\mathrm{df}_t(w)$ counts historical blocks containing $w$. The weighting emphasizes explicit references to rare file names, functions, and error symbols. We combine this cue with stored importance and the node annotations:
\begin{equation}
s_v=\mathrm{z}(u_t(v))+\lambda\,\mathrm{z}(\ell_v)
+\epsilon\,\xi_{t,v}\,\mathbf{1}[v\notin\mathcal{O}_t]
-\kappa\,\mathbf{1}[v\in\mathcal{S}_t].
\label{eq:score}
\end{equation}
Here $\mathrm{z}$ standardizes each channel over $V_t$ and $\lambda$ controls the query contribution; never-observed nodes receive a small exploration bonus ($\xi_{t,v}\in[0.5,1]$, drawn with a fixed session-and-turn seed), and stale nodes receive a penalty $\kappa$. The first two terms combine historical importance with current relevance, while the annotations adjust the priority of unmeasured and potentially superseded blocks.

\textbf{Packing historical blocks.}
Selection begins with a protected set $P_t$: all user-message nodes and the latest edit or write node per modified file, preserving task instructions and the agent's latest changes (Appendix~\ref{app:protected}). Starting from $C=P_t$, we visit the remaining nodes in descending score order, breaking ties chronologically, and add each block that fits within the nominal budget $B_t$ (\S\ref{sec:setting}). $P_t$ is retained even when its cost exceeds $B_t$. Let $C_0$ denote this initial selection.

\textbf{Including supporting context.}
For each selected block $b\in C_0$, we traverse incoming edges and add each unselected, non-stale predecessor that fits within an expanded budget of $(1+\delta)B_t$; expansion is one hop from the fixed seed set $C_0$. This couples selection across related events: a supporting block can enter through its dependency link even when its own score was insufficient for the initial packing. Appendix~\ref{app:graph-algorithm} gives the traversal order and cost bound.

\textbf{Constructing the working context.}
\label{sec:deployment}
\method{} renders the selected blocks chronologically between system instructions and the current turn. The selection stays fixed as tool steps extend the turn, allowing prefix-cache reuse, and execution feeds attention to the next update, completing the cycle.

\section{Experiments}
\label{sec:experiments}

\subsection{Experimental Setup}

\textbf{Benchmarks and models.}
We evaluate \method{} on two interactive coding benchmarks: $106$ tasks from SWE-Together~\citep{wu2026swe}, which involves repository-level tasks with tool use and user feedback, and $100$ code instances from Lost-in-Conversation (LiC)~\citep{laban2026llms}, which reveal requirements across user turns, with five conversations per instance. Our backbones are Qwen3-Coder-30B-A3B-Instruct~\citep{yang2025qwen3} and gpt-oss-20b~\citep{openai2025gptoss120bgptoss20bmodel}, served with SGLang~\citep{zheng2024sglang}. SWE-Together runs in OpenCode, and GPT-5.4 simulates all users.

\textbf{Baselines.}
We compare against full-history restoration and six compaction baselines in three groups. \emph{Heuristic} baselines select history without model computation: Recency retains the most recent content, and Random samples content and restores its order. \emph{Prompt-compression} baselines, LLMLingua-2~\citep{pan2024llmlingua} and LongLLMLingua~\citep{jiang2024longllmlingua}, compress historical text with a small model, the latter conditioned on the request. \emph{Memory-compaction} baselines use the backbone itself: Codex-style summarization~\citep{openai2025codexcli} generates a handoff summary while preserving historical user instructions, and KV eviction~\citep{liu2026practical} keeps, per attention head, the cached entries most attended by the new user message and discards the rest permanently. All compactors operate on history preceding the current user turn, with system instructions and current-turn messages retained separately; KV eviction also ranks cached system instructions.

\textbf{End-to-end protocol and budgets.}
Compaction affects subsequent responses, tool use, and user interactions, so methods develop different histories, and matching tokens at every turn would require adjusting budgets as trajectories evolve. We therefore fix each method's hyperparameters throughout execution: a nominal retention of $\beta=0.1$ for history selection, a compression rate of $0.1$ for Lingua ($0.2$ for LiC LLMLingua-2), and $10\%$ of the historical cache entries per head for KV eviction. \method{} counts protected records toward its budget and retains them when they exceed it, and summarization sets its own summary length. We compare task quality together with realized context usage and compaction overhead (Appendix~\ref{app:evaluation-details}).

\textbf{Metrics.}
We report reward on SWE-Together and accuracy on LiC, scaled by $100$, with standard deviations across repeated runs. Context usage is the mean number of tokens per call, excluding fixed system prompts, with the same accounting scope for every method. We also report the additional backbone input and output tokens used for compression, compression latency, and the memory each method keeps for compaction (Appendices~\ref{app:evaluation-details} and~\ref{app:implementation}).

\subsection{Main Results}

\input{tables/main_results}

Table~\ref{tab:main-results} reports task quality, context usage, and the additional cost of compaction. On SWE-Together, \method{} at $\beta{=}0.1$ roughly halves the historical context of full restoration while scoring $4.1$ and $0.8$ points higher with Qwen3-Coder and gpt-oss, respectively. On Lost-in-Conversation, it improves accuracy over full restoration by $19.8$ and $41.2$ points.

On LiC, \method{} exceeds Codex-style summarization by $10.0$ and $30.8$ accuracy points with mean contexts of $94$ and $73$ tokens, compared with $372$ and $602$ for summarization and $1{,}121$ and $1{,}023$ for full restoration. Summarization also processes $1{,}500$ and $1{,}971$ input tokens and generates $285$ and $591$ output tokens per invocation, whereas \method{} selects context through graph traversal and lexical matching without an additional model call.

\method{} also keeps little state for compaction: its graph of scalar scores and dependency links occupies about $19$\,KB of CPU memory per SWE-Together session with Qwen3-Coder and under $3$\,KB elsewhere. The Lingua baselines keep $2.2$ and $5.6$\,GB of compressor weights on the GPU, and KV eviction holds $30$\,MB to $1.3$\,GB of KV cache per session until the next user message arrives, because it needs that message to rank the cached entries.

Because \method{} selects from the full archive at each request, omitted blocks can return later: on SWE-Together with Qwen3-Coder, $70\%$ of turn transitions revive at least one omitted block, whereas recency and KV eviction cannot recover dropped history (Appendix~\ref{app:patterns}).

\subsection{Inference Efficiency}
\label{sec:ttft}

\begin{wrapfigure}{r}{0.5\textwidth}
  \vspace{-0.6\baselineskip}
  \centering
  \includegraphics[width=\linewidth]{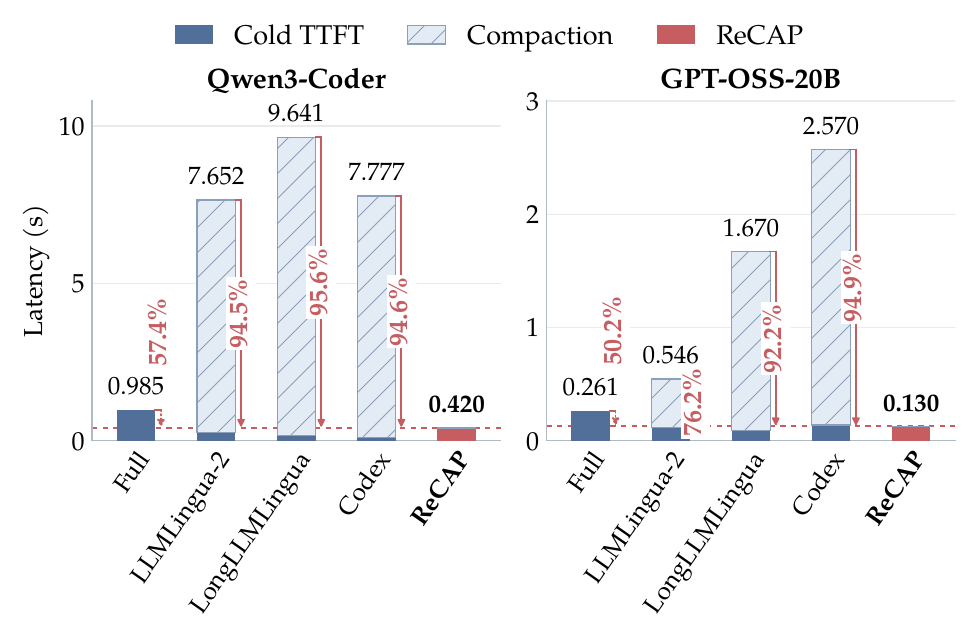}
  \vspace{-20pt}
  \caption{Estimated cold-restoration latency on SWE-Together. Bars combine TTFT and compaction overhead; arrows show reductions with \method{}. The two panels use separate scales.}
  \label{fig:swe-ttft}
  \vspace{-0.5\baselineskip}
\end{wrapfigure}
Memory compaction reduces context-encoding time, but its own computation also adds latency. We evaluate this trade-off on SWE-Together with one H100 80GB GPU, measuring cold time to first token (TTFT) for each model and method over $30$ uncached synthetic requests at its mean retained context length. Adding mean TTFT to the serial H100 compaction estimates in Table~\ref{tab:main-results} estimates the latency at a compaction boundary; \method{} uses a prepared context graph (Appendix~\ref{app:ttft-protocol}).

\method{} reduces cold TTFT from $0.985$ to $0.395$\,s for Qwen3-Coder and from $0.261$ to $0.128$\,s for gpt-oss, reductions of $59.9\%$ and $50.8\%$ over Full, or speedups of $2.49\times$ and $2.03\times$.
Compaction overhead changes the comparison among methods. The Lingua baselines achieve lower TTFT than \method{}, but their compression dominates the combined cost, and summarization adds a generation step. Including compaction, their estimated latencies range from $7.65$ to $9.64$\,s for Qwen3-Coder and from $0.55$ to $2.57$\,s for gpt-oss, and \method{} reduces these totals by $94.5$--$95.6\%$ and $76.2$--$94.9\%$ (Figure~\ref{fig:swe-ttft}). Reusing importance and dependencies from the persistent graph lets \method{} shorten the working context without an additional model call.

\subsection{Ablation Studies}
\label{sec:ablation}
\input{tables/ablation}

We ablate \method{} on the LiC code tasks with \model{Qwen3-Coder-30B-A3B} under the protocol of Table~\ref{tab:main-results}. Removing the attention graph discards importance scores, dependency expansion, and stale marks, leaving a selector that keeps the protected records and fills the remaining budget by identifier overlap with the request. The segmentation variants split plain assistant messages into paragraphs or lines instead of one block, keeping fenced code intact. All three variants score below \method{} (Table~\ref{tab:ablation}): by $8.8$ points without the attention graph, and by $2.2$ and $2.4$ points with paragraph and line blocks. They also differ in context size: without the attention graph, the working context stays at $95$ tokens on average ($94$ for \method{}), whereas finer blocks let fragments of earlier answers fit within the budget, raising it to $152$ and $168$ tokens.

\subsection{Compatibility with Prefix Caching}
\label{sec:prefix-cache}

\begin{figure}[t]
\centering
\includegraphics[width=\textwidth]{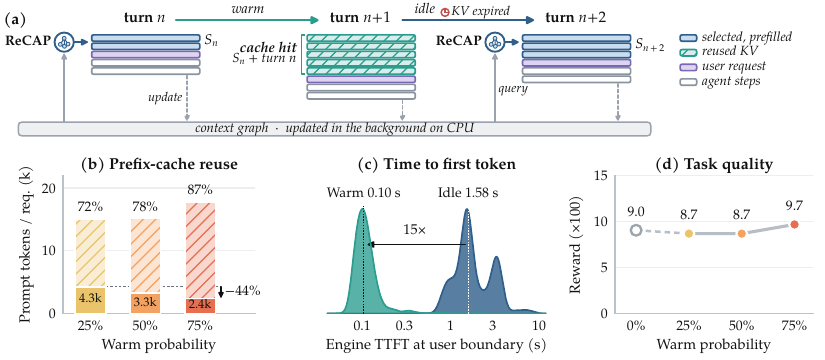}
\vspace{-15pt}
\caption{\textbf{\method{} with prefix caching} (gpt-oss-20b, SWE-Together). \textbf{(a)} Idle boundaries prefill the compact context $S_n$; warm ones reuse the cache; the graph updates in the background. \textbf{(b)} Prompt tokens per request, cached (hatched) and uncached (solid), with hit rates. \textbf{(c)} Engine TTFT at warm and idle boundaries. \textbf{(d)} Task reward versus warm probability; hollow point: all idle (Table~\ref{tab:main-results}).}
\label{fig:prefix-cache}
\end{figure}

Memory compaction targets requests that arrive after the serving cache has expired. Many user turns, however, arrive while the session's KV cache is still resident, and prefix caching already avoids re-encoding the history for them. \method{} complements prefix caching rather than replacing it: it rewrites the working context only when the cache is gone (Figure~\ref{fig:prefix-cache}a). At an idle boundary, \method{} selects a compact context $S_n$, which the engine prefills. At a warm boundary, it skips selection: the next request is appended to the resident context, so the model sees $S_n$, the complete turn $n$, and the new request, and the engine reuses their cached prefix. Within a turn, the selected history likewise stays fixed. Graph updates stay off the critical path: attention statistics are captured during inference, and the graph is updated on the CPU in the background, overlapping the next turn's generation at warm boundaries and the idle period at idle boundaries.

We evaluate this policy on SWE-Together with gpt-oss-20b served by vLLM~\citep{kwon2023efficient} with automatic prefix caching. At each user boundary, a seeded draw marks the session warm with probability $p$, keeping its cache, or idle, discarding it and running \method{}; we run $p\in\{25\%,50\%,75\%\}$ once each.
Warm boundaries reuse over $99.9\%$ of their prompt tokens from the cache. As $p$ rises from $25\%$ to $75\%$, the cache-hit rate grows from $72\%$ to $87\%$ and uncached prompt tokens per request fall by $44\%$ (Figure~\ref{fig:prefix-cache}b). The engine returns the first token in a median of $0.10$\,s at warm boundaries, compared with $1.58$\,s when an idle boundary prefills the compacted context (Figure~\ref{fig:prefix-cache}c). Task reward stays between $8.7$ and $9.7$ across the three settings, with no significant difference from the $25\%$ setting (Figure~\ref{fig:prefix-cache}d), and is $9.0$ when every boundary is idle (Table~\ref{tab:main-results}). Consecutive warm turns carry their appended history, so the average prompt grows from $15.1$k to $17.8$k tokens, but the cache serves this growth. Prefix caching thus serves turns whose cache survives, and \method{} serves turns whose cache has expired, without undoing each other.

\subsection{Attention Estimation for Black-Box Agents}
\label{sec:proxy}
\input{tables/proxy}

\method{} reads attention from the agent's own model, which a closed API does not expose. Because the graph stores block-level statistics, it can come from a different model that replays the served context with its own tokenizer and maps its attention back to the same blocks. We use Qwen3-0.6B, about $2$--$3\%$ of the agent's parameters, as the proxy for a Qwen3-Coder-30B agent from the same family and a gpt-oss-20b agent from a different family, with all other settings as in Table~\ref{tab:main-results}.

The proxy matches the agent's own attention in all three settings (Table~\ref{tab:proxy}): the differences are $+0.6$ and $-0.2$ points on LiC with gpt-oss and Qwen, and $+0.5$ points on SWE-Together with Qwen, none statistically significant. With Qwen on LiC, removing the attention graph lowers accuracy, and the proxy retains this gain, scoring $8.6$ points above the no-attention variant. A small open model can therefore supply attention for a black-box agent, including one from a different model family.

\section{Conclusion}
\label{sec:conclusion}

\method{} keeps attention that inference already computes in a persistent context graph queried at each request. Without extra model calls, it cuts compaction and cold-restoration latency by about $95\%$ versus summarization while raising LiC accuracy over full history by $19.8$ and $41.2$ points.

\subsubsection*{Acknowledgments}
The UCSB team acknowledges support from the National Science Foundation (NSF) under Grant Nos. 2338252, 2302730, and 2619240.

\bibliography{iclr2027_conference}
\bibliographystyle{iclr2027_conference}

\newpage
\appendix
\section{Failure vignettes under score-only selection}
\label{app:vignettes}

Two representative failures when selection uses scores alone, without protected records. \emph{(i) Task statement evicted at cold start:} at the first post-eviction turn of a session whose instruction was to analyze a GitHub issue (analysis only, no implementation), the selector kept the agent's own 215-character preamble and dropped the 9{,}116-character task block. The agent then asserted it had read the issue, invented a different problem (a bug in a nonexistent \texttt{reset} command), spent some twenty steps searching for files that do not exist, and shipped an unrelated edit. \emph{(ii) Own-edit record evicted:} the selector kept stale analysis prose describing the pre-edit code as current while dropping the tool-call record of the agent's own just-completed fix. On re-reading the file, the agent found code it did not remember writing, concluded that the issue had ``already been fixed'' and that no further changes were needed, and repeated that conclusion until the session ended. In both cases the lost information was recoverable from the workspace through ordinary tools---but the agent had no signal that anything needed recovering.

\section{Additional Observations and Measurement Details}
\label{app:measure}

The studies in Figure~\ref{fig:obs} use Qwen3-Coder-30B-A3B-Instruct to examine three aspects of compaction: attention concentration within a session, identifier reuse across turns, and attention redistribution under controlled requests. We describe their cohorts and read-outs separately below. We obtain attention by running the model on recorded contexts through the interface in Appendix~\ref{app:implementation}.

\subsection{Attention concentration and cross-turn reuse}
\label{app:historical-attention}

\textbf{Multi-turn attention example.}
The left panel of Figure~\ref{fig:obs} shows one coding session at six measured user-turn checkpoints. Each incoming request supplies attention queries over its preceding history. We aggregate received attention over the tokens of each historical block and normalize its mass over the history. For selection, blocks are ranked by attention mass per token and greedily packed under a $10\%$ historical-token budget. At the final checkpoint, the selected blocks contain $2{,}824$ of $28{,}309$ tokens and retain $65.01\%$ of historical attention. Heatmap columns and the concatenated context use the same token-length scale. Color denotes per-token attention divided by the maximum within each row, displayed on a cube-root scale.

\begin{figure}[t]
\centering
\includegraphics[width=\textwidth]{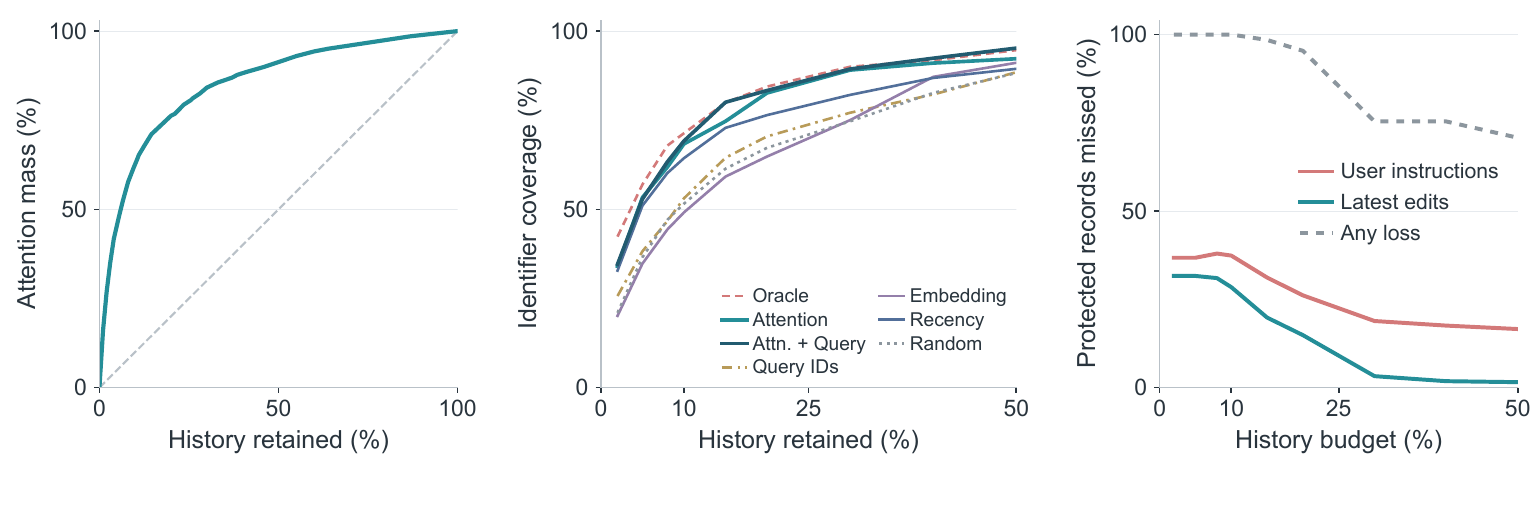}
\caption{\textbf{Additional historical-context observations.}
\textbf{Left:} Aggregate attention concentration under a retained-history budget; the diagonal denotes uniform attention per unit of history.
\textbf{Middle:} The full comparison of selection signals, including query identifiers and query embeddings.
\textbf{Right:} User-instruction and latest-edit records omitted by attention-ranked selection, together with the fraction of transitions losing any protected record. The selector receives enough budget to fit the complete protected set.}
\label{fig:obs-supplement}
\end{figure}

\textbf{Cross-turn study.}
The middle panel of Figure~\ref{fig:obs} uses $64$ transitions from eight coding-agent sessions. Past-attention scores aggregate attention sampled from the preceding completed turn across heads at layers $\{16,24,32,40\}$ of the $48$-layer model. Candidate blocks precede that completed turn. Block scores summarize received attention per token. Figure~\ref{fig:obs-supplement} also reports aggregate concentration and the full comparison of selection signals.

\textbf{Identifier-coverage proxy.}
Target identifiers are extracted from assistant and tool-result text, together with tool-call names and arguments, recorded before the following user message. The incoming user request itself is excluded from this evidence. The extractor retains code-like identifiers of at least four characters and dotted paths under a short stop list. An identifier occurring in $\mathrm{df}$ of the $n$ candidate blocks receives weight $\log(n/\mathrm{df})$. Only target identifiers appearing in the candidate history contribute to the denominator. Coverage is the fraction of their total weight present in the selected block set. Selectors rank blocks and pack them greedily under a fraction of the historical-text length; the evaluation code counts characters. The future-use oracle ranks blocks by overlap with the future target identifiers and uses the same packing rule. The combined selector adds a query-identifier score to past attention; the standalone query-identifier and embedding comparisons appear in Figure~\ref{fig:obs-supplement}.

\subsection{Controlled query study}
\label{app:query-study}

\textbf{Fixed histories and requests.}
We select one checkpoint from each of $32$ distinct coding tasks, with $16$ histories from \method{} traces and $16$ from recency-selection traces. Histories range from $8$k to $32$k rendered tokens. Each history contains explicit references to two target files, chosen before attention measurement. Historical block contents and identities remain fixed across request conditions.

For each target, we construct two short requests that ask the agent to continue or return to that file. This yields four requests per history. We repeat the comparison with longer, framed requests that explicitly ask the agent to recall earlier implementation facts, constraints, and tool observations. The two template families therefore provide eight measurements per history, or $256$ in total. Only the incoming request changes between measurements.

\textbf{Attention redistribution.}
We normalize attention mass over historical blocks after excluding the initial task instruction. For two resulting distributions $p$ and $q$, redistribution is their total-variation distance,
\begin{equation}
D_{\mathrm{TV}}(p,q)=\frac{1}{2}\sum_v |p(v)-q(v)|.
\label{eq:query-tv}
\end{equation}
Within each template family, a history contributes one paraphrase value, averaging the two same-target comparisons, and one goal-change value, averaging the four cross-target comparisons. Statistical analysis treats histories as paired observations.

\textbf{Results and visualization.}
For short requests, mean redistribution is $5.33\%$ under paraphrasing and $10.13\%$ under a target change. The paired difference is $4.81$ percentage points (task-bootstrap $95\%$ CI $[3.41,6.25]$) and is positive in $28/32$ histories. Framed requests give $3.82\%$ and $8.46\%$, respectively, with a paired difference of $4.64$ points ($[3.64,5.69]$), positive in $30/32$ histories. Confidence intervals use $10{,}000$ task-bootstrap resamples.

\begin{figure}[t]
\centering
\includegraphics[width=\textwidth]{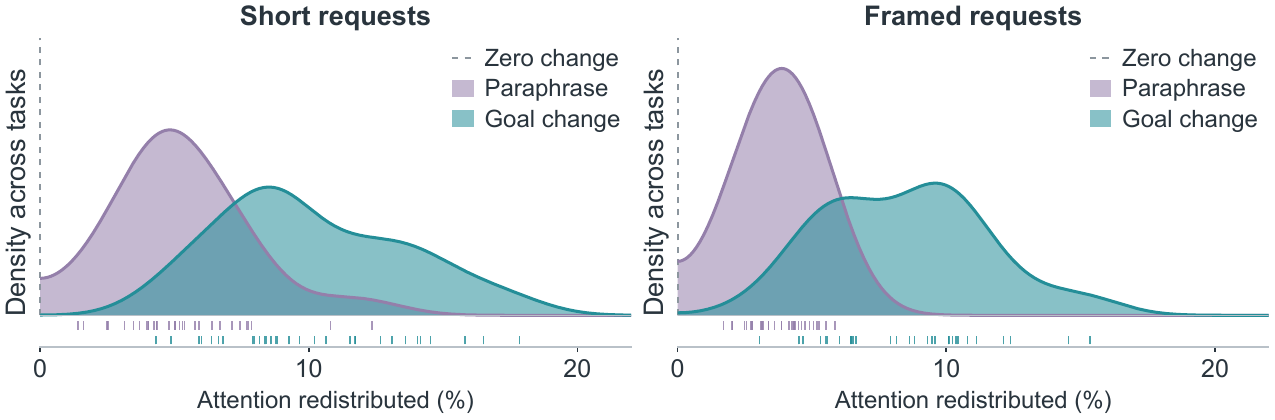}
\caption{\textbf{Query effects across request templates.} Paraphrases of the same target and requests about different targets are compared on the same $32$ histories. Longer framed requests explicitly ask the agent to recall earlier facts and constraints. Both template families show greater attention redistribution under a target change. Curves share a smoothing bandwidth; bottom ticks identify the paired task-level observations, and the dashed line marks zero redistribution.}
\label{fig:query-robustness}
\end{figure}

\subsection{Preserving instructions and recent modifications}
\label{app:protected}

User instructions specify the task, and the latest edit or write record for each modified file records the agent's most recent changes. We protect these blocks as the session's skeleton. In the cross-turn corpus, this set occupies a median $10.3\%$ of historical text. To measure whether attention ranking retains it, we give the selector budget $\max(\beta\,c(V_t),c(P_t))$, so the complete skeleton fits. At $\beta=0.1$, score-only selection still omits approximately $37\%$ of user messages and $28\%$ of latest-edit records, and every transition loses at least one protected block (Figure~\ref{fig:obs-supplement}, right). These measurements motivate explicit protection during packing. Appendix~\ref{app:vignettes} gives two corresponding failure examples.

\section{Evaluation protocol and cost accounting}
\label{app:evaluation-details}

\textbf{Tasks and budgets.}
Table~\ref{tab:main-results} reports SWE-Together results on $106$ tasks and Lost-in-Conversation results on $100$ instances, with five conversations scheduled per instance. For LiC, the $k$th conversation of every task forms repeat $k$. We report the mean and sample standard deviation ($n-1$ denominator) of the five repeat accuracies. Each repeat has $100$ tasks, with missing or unscored conversations counted as unsuccessful. SWE-Together standard deviations summarize two replicates.

For history selection, the budget is computed from each method's own accumulated history. Lingua receives its compression rate through the compressor's own interface. LiC Recency and Random use line-level splitting of assistant text, while \method{} uses message-level blocks. \method{}'s protected blocks consume the nominal allowance before other blocks are added, and dependency expansion allows up to $10\%$ additional budget, subject to the bound in Eq.~\ref{eq:retention-bound}. These rules and the independently generated histories determine realized usage. The Full row summarizes the full-restoration policy's own trajectories; per-request compression ratios use each compactor's own history before and after compression.

\textbf{Token and latency accounting.}
For SWE-Together, \emph{Ctx.} reports the mean retained historical content per request. Every method excludes the original system instructions, tool definitions, and current turn from this quantity. For LiC, \emph{Ctx.} subtracts the fixed rendered system prefix from each complete input: $73$ tokens for Qwen3-Coder and $135$ for gpt-oss, the same for all methods. The remaining count includes retained history, the current user turn, message framing, and any omission notice. LiC means cover all completed assistant calls, including the first.

Compressor prompt and completion counts use serving-engine usage; a reused summary incurs no new compression call. LLMLingua uses a separate small model, whose computation appears in compression latency. Compression latencies of the baselines are measured on one H100, one request at a time (Appendix~\ref{app:ttft-protocol}). For LiC, latency is the mean over $30$ sampled compression invocations after $10$ warm-up invocations, timed until the full compressed output is returned. \method{}'s selection runs on the CPU without a model call; its \emph{+Lat.} is the mean selection time on one CPU core (Appendix~\ref{app:selection-cost}).

\textbf{KV eviction.}
At the start of each user turn after the first, it scores every resident historical entry by its attention from the new user message and keeps $\lceil 0.1N\rceil$ entries per head, where $N$ counts all historical tokens, including those evicted earlier. The current input is kept in full, and evicted entries are never recomputed. After eviction, the engine recomputes the final input token against the compacted cache; we omit this single token from \emph{+Pre.} Its \emph{+Lat.} is the mean over $30$ compaction events sampled from the quality runs and replayed on one H100 with caches of the recorded sizes, one event at a time after $10$ warm-up events. Each measurement covers scoring, selection, the copy into compacted tensors, and the one-token recomputation. Scoring, selection, and copying take $5$--$32$\,ms on average, and the one-token recomputation takes the rest. For \emph{Ctx.}, we count the historical entries retained per head, adding the current input on LiC. These entries may include cached system instructions, so the value is an upper bound on retained history under the other methods' accounting.

\textbf{Memory accounting.}
\emph{+Mem.} counts memory that a method keeps for compaction, beyond the backbone weights and the KV cache of the working context. For the Lingua baselines, it is the compressor weights: $2.24$\,GB for LLMLingua-2 ($559$M parameters in FP32) and $5.56$\,GB for LongLLMLingua (Phi-2, $2.78$B parameters in FP16), excluding activations. For KV eviction, it is the mean KV cache a session holds when the next user message arrives, before eviction. For \method{}, it is the context graph at the end of each session, averaged over sessions, stored with $12$ bytes per node (identifier, score, and flags) and $12$ bytes per edge (two identifiers and a weight). The full history itself is kept by the agent harness for every method and is not counted. We count at most $6.5$ edges per block, the rate measured on gpt-oss SWE-Together sessions. Because an edge requires more than $\theta{=}0.02$ of the dependent block's attention mass, each block has at most $49$ incoming edges; even this worst case stays below $0.8$\,MB per session. Full restoration, Recency, Random, and summarization keep no additional state.

\subsection{Cold-TTFT measurement protocol}
\label{app:ttft-protocol}

We measure TTFT on one NVIDIA H100 80GB HBM3 GPU with SGLang 0.5.17, PyTorch 2.11.0 with CUDA 13.0, Transformers 5.12.1, and FA3 attention. Qwen3-Coder uses native BF16 weights and gpt-oss uses native MXFP4 weights. Each condition contains $30$ timed synthetic requests after warm-up. No request reuses cached tokens.

Synthetic input lengths match the SWE-Together mean historical-context sizes in Table~\ref{tab:main-results}. In the order Full, LLMLingua-2, LongLLMLingua, Summarize (Codex), and \method{}, these lengths are $28{,}504$, $11{,}062$, $7{,}429$, $5{,}239$, and $15{,}344$ tokens for Qwen3-Coder, and $8{,}508$, $3{,}340$, $2{,}416$, $4{,}543$, and $4{,}043$ tokens for gpt-oss. These requests measure cold TTFT at historical-context lengths, excluding the separately supplied system prompt, tool definitions, and current turn.

SWE-Together compaction costs are measured on the same GPU, one request at a time: the Lingua compressor shares the GPU with the serving engine, and summarization is a complete additional model call, including prefill and generation, on an idle engine with its cache cleared.

For each condition, Figure~\ref{fig:swe-ttft} adds the mean measured TTFT to the compaction latency in the main table, estimating a request that performs compaction before cold restoration. The reductions shown by the arrows are $100(1-L_{\method{}}/L_{\mathrm{baseline}})$, where $L$ is this combined estimate.

\subsection{Compaction pattern analysis}
\label{app:patterns}
\label{sec:patterns}

\begin{figure}[ht]
  \centering
  \begin{minipage}[c]{0.5\textwidth}
    \includegraphics[width=\linewidth]{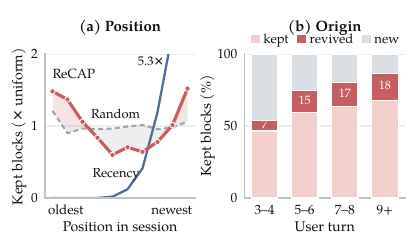}
  \end{minipage}\hfill
  \begin{minipage}[c]{0.47\textwidth}
  \caption{History retained by \method{} on SWE-Together with Qwen3-Coder.
  \textbf{(a)} Retained blocks in each tenth of the session, relative to a uniform spread; counts include protected records.
  \textbf{(b)} Origin of the blocks retained at each user turn: kept from the previous working context, revived after being omitted from it, or created since.}
  \label{fig:patterns}
  \end{minipage}
\end{figure}

We examine which history \method{} carries into the working context, using the compaction records of the Qwen3-Coder SWE-Together runs in Table~\ref{tab:main-results}. Figure~\ref{fig:patterns}a divides each session into ten equal spans by block position and compares each span's share of retained blocks with a uniform spread. Recency concentrates on the latest span, at $5.3\times$ its uniform share, and retains almost nothing from the oldest $40\%$ of the session, while Random stays close to uniform. \method{} draws from both ends: the oldest and newest spans each receive about $1.5\times$ their uniform share, and the middle falls to $0.6\times$. The oldest $30\%$ of the session supplies $39\%$ of the blocks \method{} retains, compared with $31\%$ for random selection and none for recency.

Because \method{} selects from the full archive at each request, blocks omitted from one working context can return in a later one (Figure~\ref{fig:patterns}b). About $60\%$ of retained blocks carry over from the previous working context, and the share of revived blocks grows from $7\%$ in turns 3--4 to $18\%$ from turn 9 onward; $70\%$ of turn transitions revive at least one omitted block. Recency never revives a block once its window has passed it, and KV compaction discards evicted entries permanently, so a later request cannot recover them without re-encoding the full history~\citep{liu2026practical,zweiger2026fast}.

\subsection{Example of a revived block}
\label{app:revival}

The example below comes from a SWE-Together session with Qwen3-Coder. At the first user turn, the agent checked the authentication-event routes and handlers with \texttt{grep}. The resulting block is not protected and had no recorded importance, so it stayed outside the working context for the next seven user turns. At turn 8, the user asks the agent to verify these handlers and names six of the functions that appear in the block. These rare identifiers give the block the highest query-overlap score among the $98$ history blocks, and \method{} selects it back into the working context under the same $10\%$ budget. Because selection runs over the full history at each request, content omitted for several turns can return once a request refers to it.

\begin{tcolorbox}[colback=BlueGreen!6, colframe=BlueGreen!70!black, boxrule=0.6pt, arc=2pt,
  left=6pt, right=6pt, top=4pt, bottom=4pt, fonttitle=\bfseries\small,
  title={History block from user turn 0 (omitted at turns 1--7, selected at turn 8)}]
\small
Let me verify that all the necessary endpoints have been added by running a quick check:\\[2pt]
{\footnotesize\ttfamily
\$ cd /workspace/hyperswitch \&\& grep -n "get\_.*auth\_event" crates/router/src/analytics.rs\\
318: .route(web::post().to(get\_org\_auth\_event\_metrics)),\\
409: .route(web::post().to(get\_profile\_auth\_event\_metrics)),\\
{[}...{]}\\
1121: pub async fn get\_org\_auth\_event\_metrics(\\
1166: pub async fn get\_profile\_auth\_event\_metrics(\\
{[}...{]}\\
3087: pub async fn get\_profile\_auth\_event\_sankey(}
\end{tcolorbox}

\begin{tcolorbox}[colback=RedOrange!5, colframe=RedOrange!70!black, boxrule=0.6pt, arc=2pt,
  left=6pt, right=6pt, top=4pt, bottom=4pt, fonttitle=\bfseries\small,
  title={User request at turn 8}]
\small
wait, this still looks incomplete/wrong --- your final diff only shows route additions, but earlier you said you added new handler fns and core auth changes too. pls verify the new handlers actually exist and are wired, [...] plus a quick grep for {\footnotesize\ttfamily get\_org\_auth\_event\_metrics|\allowbreak{}get\_profile\_auth\_event\_metrics|\allowbreak{}get\_org\_auth\_event\_sankey|\allowbreak{}get\_profile\_auth\_event\_sankey|\allowbreak{}get\_org\_auth\_events\_filters|\allowbreak{}get\_profile\_auth\_events\_filters}, and fix/commit if anything's missing.
\end{tcolorbox}

\subsection{Selection cost}
\label{app:selection-cost}

We time \method{}'s context selection, from block segmentation through scoring, packing, and dependency expansion, on one CPU core (AMD EPYC 7453) for requests of the sizes in Table~\ref{tab:main-results}. Table~\ref{tab:selcost} reports the per-request time. Selection takes about a millisecond on LiC and on gpt-oss SWE-Together sessions and a median of $20$\,ms on the longer Qwen3-Coder SWE-Together sessions, well below the compression latency of every baseline. Almost all of this time extracts identifiers from the history text, so it grows linearly with history length; because blocks do not change once written, their identifier sets can be computed once and reused. Updating the graph from attention statistics, including the moving average, edge merging, and the supersession test, takes a median of $3.7$\,ms per turn with Qwen3-Coder and $0.35$\,ms with gpt-oss on SWE-Together.

\begin{table}[h]
\centering
\small
\caption{Per-request selection time of \method{} on one CPU core.}
\label{tab:selcost}
\begin{tabular}{llcccc}
\toprule
Benchmark & Backbone & Blocks (median) & Median (ms) & P90 (ms) & Max (ms) \\
\midrule
SWE-Together & Qwen3-Coder & 132 & 20.2 & 45.2 & 112.6 \\
SWE-Together & gpt-oss & 18 & 1.4 & 4.4 & 20.3 \\
LiC & Qwen3-Coder & 6 & 0.30 & 0.80 & 1.69 \\
LiC & gpt-oss & 6 & 0.26 & 0.65 & 1.51 \\
\bottomrule
\end{tabular}
\end{table}

\subsection{Summarization baseline prompts}
\label{app:codex-prompt}

The Codex-style summarization baseline uses the compaction prompts of the Codex CLI~\citep{openai2025codexcli} verbatim. The instruction below is appended to the history as a final user message, and the returned summary follows the prefix in the compacted context.

\begin{tcolorbox}[colback=BlueGreen!6, colframe=BlueGreen!70!black, boxrule=0.6pt, arc=2pt,
  left=6pt, right=6pt, top=4pt, bottom=4pt, fonttitle=\bfseries\small,
  title={Compaction instruction (final user message)}]
\small
You are performing a CONTEXT CHECKPOINT COMPACTION. Create a handoff summary for another LLM that will resume the task.

\smallskip
Include:
\begin{itemize}[leftmargin=12pt, nosep]
\item Current progress and key decisions made
\item Important context, constraints, or user preferences
\item What remains to be done (clear next steps)
\item Any critical data, examples, or references needed to continue
\end{itemize}

\smallskip
Be concise, structured, and focused on helping the next LLM seamlessly continue the work.
\end{tcolorbox}

\begin{tcolorbox}[colback=RedOrange!5, colframe=RedOrange!70!black, boxrule=0.6pt, arc=2pt,
  left=6pt, right=6pt, top=4pt, bottom=4pt, fonttitle=\bfseries\small,
  title={Summary prefix (precedes the summary in the compacted context)}]
\small
Another language model started to solve this problem and produced a summary of its thinking process. You also have access to the state of the tools that were used by that language model. Use this to build on the work that has already been done and avoid duplicating work. Here is the summary produced by the other language model, use the information in this summary to assist with your own analysis:
\end{tcolorbox}

\section{Implementation details and hyperparameters}
\label{app:method-details}

\subsection{Attention access and serving realization}
\label{app:implementation}

\textbf{Attention interface.}
Native integration obtains sampled attention rows from retained query vectors, or their source hidden states, and resident keys. Importance extraction samples current-turn positions across a layer band; dependency extraction samples positions within each block at one selected layer. The sampled rows are reduced to block statistics at the turn boundary. Subsequent context selection reads these stored statistics and the graph's adjacency information.

\textbf{Request construction and context limits.}
The proxy rewrites the request's message array with the selected blocks. When history is omitted, a system notice identifies the omission and states that the workspace remains available for retrieving further details. Selection uses the same history boundary, graph version, request, and random seed throughout a user turn. The engine enforces a 360k-character request cap by dropping the oldest selected blocks when the full request exceeds the limit. This guardrail takes precedence over protected-record retention and prefix stability.

\subsection{Graph construction and selection}
\label{app:graph-algorithm}

\textbf{Graph state and initialization.}
The persistent state contains the graph $G_t=(V_t,E_t)$ with node scores $u_t$ and edge weights $w_t$, and the node annotations $\mathcal{O}_t$ and $\mathcal{S}_t$. Node scores start at zero; edges and annotation sets start empty. A graph update registers newly observed blocks and revises scores only for $R_t$. Edges, stale marks, and observation membership accumulate across updates. The state requires $O(|V_t|+|E_t|)$ scalar metadata in addition to the history text. The protected set $P_t$ is recomputed from message roles and edit/write records for each new turn.

\textbf{Dependency and supersession rules.}
For sampled positions $W_t(b)$ within block $b$, the strength of an edge from $a$ to $b$ is
\begin{equation}
d_t(a,b)=\frac{1}{|W_t(b)|}\sum_{i\in W_t(b)}
\sum_{j\in\mathrm{span}_t(a)}\bar a_t^{\mathrm{dep}}(i\to j),
\label{eq:dependency}
\end{equation}
where $\bar a_t^{\mathrm{dep}}$ averages heads at the dependency layer. We add an edge when $a$ precedes $b$ and $d_t(a,b)>\theta$, retaining its maximum measured strength across updates. Expansion uses the retained edge's presence. Supersession compares observed pairs in chronological order. An earlier block $a$ is marked stale when its identifier-set Jaccard similarity with a later block $b$ exceeds $0.5$ and $\bar\rho_t(b)>2\bar\rho_t(a)+0.05\bar\rho_t^{\max}$. The identifiers for this test are extracted from truncated block text, as specified below. Stale marks persist; selection applies their score penalty and skips stale predecessors during expansion.

\textbf{Packing order and budget.}
Algorithm~\ref{alg:recap} summarizes the two graph operations. Ranked packing breaks score ties by chronological order. Expansion visits the fixed seed set $C_0$ in chronological order and each seed's edges in their stored order. Newly added predecessors are not used as expansion seeds. Before the serving cap is applied, the selected set satisfies
\begin{equation}
P_t\subseteq C,\qquad c(C)\le\max\{c(P_t),(1+\delta)B_t\}.
\label{eq:retention-bound}
\end{equation}
If protected records already exceed $(1+\delta)B_t$, the packing and expansion tests add no further blocks. System instructions, the omission notice, and the current turn are outside this historical-block budget.

\begin{algorithm}[t]
\caption{Updating and querying the persistent context graph.}
\label{alg:recap}
\begin{algorithmic}[1]
\STATE \textbf{procedure} \textsc{UpdateGraph}$(R_t,\,\text{sampled attention},\,G,\,\mathcal{O},\,\mathcal{S})$
\STATE \quad register newly observed blocks in $V$ with zero initial importance
\STATE \quad compute $\bar\rho_t(v)$ and update $u(v)$ for $v\in R_t$ using Eq.~\ref{eq:utility}
\STATE \quad $\mathcal{O}\leftarrow\mathcal{O}\cup R_t$
\STATE \quad add edges $(a,b)$ with $a<b$ and $d_t(a,b)>\theta$ using Eq.~\ref{eq:dependency}
\STATE \quad set each edge weight $w(a,b)$ to its maximum observed $d_t(a,b)$
\STATE \quad mark $a\in\mathcal{S}$ when a later $b$ passes the supersession test
\STATE \textbf{procedure} \textsc{SelectContext}$(H_t,\,q_{t+1},\,G_t,\,\mathcal{O}_t,\,\mathcal{S}_t)$
\STATE \quad register any remaining blocks in $H_t$; derive $P_t$ and $B_t=\lfloor\beta c(V_t)\rfloor$
\STATE \quad score nodes using Eq.~\ref{eq:score}; $C\leftarrow P_t$
\STATE \quad \textbf{for} $v\in V_t\setminus P_t$ in descending score order:
\STATE \qquad \textbf{if} $c(C)+c(v)\le B_t$: $C\leftarrow C\cup\{v\}$
\STATE \quad $C_0\leftarrow C$ \hfill $\triangleright$ fixed seeds for one-hop expansion
\STATE \quad \textbf{for} $b\in C_0$ in chronological order, each predecessor $a$ in stored order:
\STATE \qquad \textbf{if} $a\notin C\cup\mathcal{S}_t$ and $c(C)+c(a)\le(1+\delta)B_t$: $C\leftarrow C\cup\{a\}$
\STATE \quad \textbf{return} selected blocks in chronological order
\end{algorithmic}
\end{algorithm}

\subsection{Hyperparameters and preprocessing}

Table~\ref{tab:hparams} lists the graph-update and compaction settings. Character cost counts message content and serialized tool-call objects. Identifier extraction for Eq.~\ref{eq:lexical} uses message content and matches code-like tokens of length $\ge4$, including dotted file paths, under a short stop list; document frequencies are computed within the session. Supersession compares identifier sets from up to 4,000 characters of block content and at most 400 matches. Received attention is clipped at its 99.9th percentile before block averaging. A zero maximum in importance normalization contributes a zero observation; standardization returns zero for a constant channel. Before any attention is observed, query overlap and the exploration bonus determine the unprotected ranking.

\begin{table}[h]
\centering
\caption{\method{} hyperparameters. Selection and update thresholds are shared across tasks and replicates; attention layers depend on the model architecture.}
\label{tab:hparams}
\begin{tabular}{llc}
\toprule
Symbol & Meaning & Value \\
\midrule
$\beta$ & nominal history-character retention fraction & $0.10$ \\
$\lambda$ & weight of the lexical channel in Eq.~\ref{eq:score} & $0.5$ \\
$\alpha$ & importance EMA rate (\S\ref{sec:feedback}) & $0.6$ \\
$\epsilon$ & exploration bonus scale for unobserved blocks & $0.1$ \\
$\kappa$ & stale penalty (in standardized-score units) & $2.0$ \\
$\delta$ & allowance for one-hop dependency expansion & $0.1$ \\
$L$ & Qwen feedback layers (48 layers, zero-indexed) & $\{16, 24, 32, 40\}$ \\
$L$ & gpt-oss feedback layers (24 layers, zero-indexed) & $\{11, 15, 19, 23\}$ \\
$|W_t|$ & sampled current-turn positions per feedback pass & $\le 640$ \\
--- & sampled positions per block for edge extraction & $\le 96$ \\
$\theta$ & attention-mass threshold for a dependency edge & $0.02$ \\
--- & supersede identifier-Jaccard threshold & $0.5$ \\
--- & supersede attention dominance & $\bar\rho_t(b) > 2\,\bar\rho_t(a) + 0.05\,\bar\rho_t^{\max}$ \\
--- & received-attention tail clip & $99.9$th percentile \\
\bottomrule
\end{tabular}
\end{table}

\end{document}

%% file: math_commands.tex
\usepackage{amsmath,amsfonts,bm}

\def\eqref#1{equation~\ref{#1}}

\def\1{\bm{1}}

\DeclareMathAlphabet{\mathsfit}{\encodingdefault}{\sfdefault}{m}{sl}
\SetMathAlphabet{\mathsfit}{bold}{\encodingdefault}{\sfdefault}{bx}{n}

%% file: tables/main_results.tex
\providecommand{\std}[1]{{\scriptsize\color{gray}$\pm$#1}}      %
\providecommand{\best}[1]{\textbf{#1}}                            %
\providecommand{\second}[1]{\underline{#1}}                       %
\providecommand{\model}[1]{\texttt{\small #1}}                    %
\providecommand{\red}[1]{\textcolor{RedOrange}{$\uparrow$#1}}     %
\providecommand{\blue}[1]{\textcolor{BlueGreen}{$\downarrow$#1}}  %
\providecommand{\zero}[1]{\textcolor{gray}{#1}}                   %
\providecommand{\ci}[1]{{\scriptsize\color{gray}[#1]}}            %
\providecommand{\grp}[1]{\multirow{2}{*}{\scshape\makecell{#1}}} %
\providecommand{\bench}[1]{\multicolumn{16}{c}{\textit{#1}}\\}

\begin{table}[t]
  \centering
  \setlength{\abovecaptionskip}{0pt}
  \small
  \setlength{\tabcolsep}{3.2pt}
  \renewcommand{\arraystretch}{1.12}
  \caption{Main results on SWE-Together (reward) and Lost-in-Conversation (LiC, accuracy), scaled by 100; $\pm$: standard deviation across runs.
  $\Delta$ is \method{}'s score minus the row's (\textcolor{RedOrange}{orange}: \method{} higher; \textcolor{BlueGreen}{teal}: lower).
  \emph{Ctx.}: mean tokens per call, excluding fixed system prompts.
  Compaction adds, per invocation, \emph{+Pre.} prompt and \emph{+Dec.} completion tokens on the backbone (means) and \emph{+Lat.} seconds (one H100; \method{} on CPU);
  \emph{+Mem.}: memory it keeps in MB (GPU for compressors and KV cache, CPU for the context graph).
  \textbf{Bold}/\underline{underline}: best/second-best scores; Appendix~\ref{app:evaluation-details} gives the accounting.}
  \label{tab:main-results}
  \resizebox{\linewidth}{!}{
  \begin{tabular}{llcccccccccccccc}
    \toprule
    \multicolumn{2}{c}{\multirow{2.3}{*}{\textbf{Method}}}
    & \multicolumn{7}{c}{\model{Qwen3-Coder-30B-A3B}}
    & \multicolumn{7}{c}{\model{gpt-oss-20b}} \\
    \cmidrule(lr){3-9} \cmidrule(lr){10-16}
    &
    & Score$\uparrow$ & $\Delta$ & Ctx.$\downarrow$ & +Pre.$\downarrow$ & +Dec.$\downarrow$ & +Lat.$\downarrow$ & +Mem.$\downarrow$
    & Score$\uparrow$ & $\Delta$ & Ctx.$\downarrow$ & +Pre.$\downarrow$ & +Dec.$\downarrow$ & +Lat.$\downarrow$ & +Mem.$\downarrow$ \\
    \midrule
    \bench{SWE-Together}
    \midrule
    \textsc{No compr.} & Full
    & 24.4\std{1.87} & \red{4.1} & 28,504 & 0 & 0 & 0 & 0
    & 8.2\std{0.15} & \red{0.8} & 8,508 & 0 & 0 & 0 & 0 \\
    \midrule
    \grp{Heuristic}
    & Recency
    & 26.6\std{0.40} & \red{1.9} & 10,154 & 0 & 0 & 0 & 0
    & 8.4\std{0.08} & \red{0.6} & 1,264 & 0 & 0 & 0 & 0 \\
    & Random
    & 23.8\std{0.99} & \red{4.7} & 12,712 & 0 & 0 & 0 & 0
    & 7.4\std{0.35} & \red{1.6} & 1,011 & 0 & 0 & 0 & 0 \\
    \midrule
    \grp{Prompt\\compr.}
    & LLMLingua-2
    & 23.9\std{1.17} & \red{4.6} & 11,062 & 0 & 0 & 7.39 & 2,236
    & 7.8\std{0.22} & \red{1.2} & 3,340 & 0 & 0 & 0.43 & 2,236 \\
    & LongLLMLingua
    & 26.6\std{0.76} & \red{1.9} & 7,429 & 0 & 0 & 9.48 & 5,559
    & 7.9\std{0.04} & \red{1.1} & 2,416 & 0 & 0 & 1.58 & 5,559 \\
    \midrule
    \grp{Memory\\compact.}
    & Summarize
    & \best{28.7}\std{1.26} & \blue{0.2} & 5,239 & 71,530 & 397 & 7.67 & 0
    & 8.1\std{0.45} & \red{0.9} & 4,543 & 7,664 & 678 & 2.43 & 0 \\
    & KV eviction
    & 24.1\std{0.54} & \red{4.4} & 4,706 & 0 & 0 & 0.41 & 1,292
    & \second{8.5}\std{0.30} & \red{0.5} & 1,638 & 0 & 0 & 0.29 & 255 \\
    \midrule
    \rowcolor{gray!10}
    \cellcolor{white}\textsc{Ours} & \method{}
    & \second{28.5}\std{0.40} & --- & 15,344 & 0 & 0 & 0.02 & 0.02
    & \best{9.0}\std{0.01} & --- & 4,043 & 0 & 0 & $<$0.01 & $<$0.01 \\
    \midrule
    \bench{Lost-in-Conversation}
    \midrule
    \textsc{No compr.} & Full
    & 65.6\std{1.82} & \red{19.8} & 1,121 & 0 & 0 & 0 & 0
    & 26.4\std{3.44} & \red{41.2} & 1,023 & 0 & 0 & 0 & 0 \\
    \midrule
    \grp{Heuristic}
    & Recency
    & 59.0\std{2.83} & \red{26.4} & 248 & 0 & 0 & 0 & 0
    & 22.6\std{2.41} & \red{45.0} & 174 & 0 & 0 & 0 & 0 \\
    & Random
    & 75.2\std{3.11} & \red{10.2} & 170 & 0 & 0 & 0 & 0
    & 33.2\std{2.86} & \red{34.4} & 158 & 0 & 0 & 0 & 0 \\
    \midrule
    \grp{Prompt\\compr.}
    & LLMLingua-2
    & 48.8\std{3.96} & \red{36.6} & 253 & 0 & 0 & 1.20 & 2,236
    & 19.2\std{5.07} & \red{48.4} & 17 & 0 & 0 & 0.94 & 2,236 \\
    & LongLLMLingua
    & 65.6\std{4.56} & \red{19.8} & 163 & 0 & 0 & 1.26 & 5,559
    & 21.0\std{2.83} & \red{46.6} & 17 & 0 & 0 & 1.09 & 5,559 \\
    \midrule
    \grp{Memory\\compact.}
    & Summarize
    & \second{75.4}\std{2.70} & \red{10.0} & 372 & 1,500 & 285 & 1.45 & 0
    & \second{36.8}\std{5.12} & \red{30.8} & 602 & 1,971 & 591 & 2.38 & 0 \\
    & KV eviction
    & 70.0\std{3.58} & \red{15.4} & 146 & 0 & 0 & 0.38 & 62
    & 33.4\std{2.87} & \red{34.2} & 256 & 0 & 0 & 0.31 & 30 \\
    \midrule
    \rowcolor{gray!10}
    \cellcolor{white}\textsc{Ours} & \method{}
    & \best{85.4}\std{1.95} & --- & 94 & 0 & 0 & $<$0.01 & $<$0.01
    & \best{67.6}\std{3.58} & --- & 73 & 0 & 0 & $<$0.01 & $<$0.01 \\
    \bottomrule
  \end{tabular}
  }
  \vspace{-8pt}
\end{table}

%% file: tables/ablation.tex
\providecommand{\std}[1]{{\scriptsize\color{gray}$\pm$#1}}
\providecommand{\red}[1]{\textcolor{RedOrange}{$\uparrow$#1}}

\begin{wraptable}{r}{0.5\textwidth}
  \vspace{-1.0\baselineskip}
  \centering
  \small
  \setlength{\tabcolsep}{4.0pt}
  \renewcommand{\arraystretch}{1.15}
  \caption{Component ablations on Lost-in-Conversation with \model{Qwen3-Coder-30B-A3B} ($\beta{=}0.1$).
  $\Delta$ is \method{}'s accuracy minus the variant's.
  \emph{Ctx.} is the mean context size in tokens, excluding the fixed system prompt, as in Table~\ref{tab:main-results}.}
  \label{tab:ablation}
  \vspace{3pt}
  \begin{tabular}{lccc}
    \toprule
    \textbf{Variant} & Acc.$\uparrow$ & $\Delta$ & Ctx.$\downarrow$ \\
    \midrule
    \rowcolor{gray!10}
    \method{} & \textbf{85.4}\std{1.95} & --- & 94 \\
    \midrule
    w/o attention graph & 76.6\std{1.95} & \red{8.8} & 95 \\
    \midrule
    paragraph blocks & 83.2\std{1.92} & \red{2.2} & 152 \\
    line blocks & 83.0\std{2.45} & \red{2.4} & 168 \\
    \bottomrule
  \end{tabular}
  \vspace{-0.8\baselineskip}
\end{wraptable}

%% file: tables/proxy.tex
\providecommand{\std}[1]{{\scriptsize\color{gray}$\pm$#1}}

\begin{wraptable}{r}{0.5\textwidth}
  \vspace{-1.0\baselineskip}
  \centering
  \small
  \setlength{\tabcolsep}{3.5pt}
  \renewcommand{\arraystretch}{1.15}
  \caption{\method{} with attention from the agent or a Qwen3-0.6B proxy ($\beta{=}0.1$).
  Scores are scaled by 100; $\pm$: standard deviation across runs.}
  \label{tab:proxy}
  \vspace{3pt}
  \begin{tabular}{lccc}
    \toprule
    & \multicolumn{2}{c}{LiC} & SWE \\
    \cmidrule(lr){2-3} \cmidrule(lr){4-4}
    \textbf{Attention} & gpt-oss & Qwen & Qwen \\
    \midrule
    \rowcolor{gray!10}
    agent's own & 67.6\std{3.58} & 85.4\std{1.95} & 28.5\std{0.40} \\
    Qwen3-0.6B proxy & 68.2\std{3.49} & 85.2\std{2.17} & 29.0\std{0.64} \\
    \midrule
    none & 63.4\std{2.61} & 76.6\std{1.95} &  25.4\std{0.22}\\
    \bottomrule
  \end{tabular}
  \vspace{-0.8\baselineskip}
\end{wraptable}

%% file: iclr2027_conference.bbl
\begin{thebibliography}{29}
\providecommand{\natexlab}[1]{#1}
\providecommand{\url}[1]{\texttt{#1}}
\expandafter\ifx\csname urlstyle\endcsname\relax
  \providecommand{\doi}[1]{doi: #1}\else
  \providecommand{\doi}{doi: \begingroup \urlstyle{rm}\Url}\fi

\bibitem[Feng et~al.(2026)Feng, Lv, Cao, Xie, and Zhou]{feng2026ada}
Yuan Feng, Junlin Lv, Yukun Cao, Xike Xie, and S~Kevin Zhou.
\newblock Ada-kv: Optimizing kv cache eviction by adaptive budget allocation for efficient llm inference.
\newblock \emph{Advances in Neural Information Processing Systems}, 38:\penalty0 113152--113188, 2026.

\bibitem[Gao et~al.(2024)Gao, He, Sharma, Kang, Jevdjic, Deng, Yang, Yu, and Zuo]{gao2024cost}
Bin Gao, Zhuomin He, Puru Sharma, Qingxuan Kang, Djordje Jevdjic, Junbo Deng, Xingkun Yang, Zhou Yu, and Pengfei Zuo.
\newblock $\{$Cost-Efficient$\}$ large language model serving for multi-turn conversations with $\{$CachedAttention$\}$.
\newblock In \emph{2024 USENIX annual technical conference (USENIX ATC 24)}, pp.\  111--126, 2024.

\bibitem[Gao et~al.(2025)Gao, Chen, and Shu]{gao2025fast}
Shiwei Gao, Youmin Chen, and Jiwu Shu.
\newblock Fast state restoration in llm serving with hcache.
\newblock In \emph{Proceedings of the Twentieth European Conference on Computer Systems}, pp.\  128--143, 2025.

\bibitem[Jiang et~al.(2024)Jiang, Wu, Luo, Li, Lin, Yang, and Qiu]{jiang2024longllmlingua}
Huiqiang Jiang, Qianhui Wu, Xufang Luo, Dongsheng Li, Chin-Yew Lin, Yuqing Yang, and Lili Qiu.
\newblock Longllmlingua: Accelerating and enhancing llms in long context scenarios via prompt compression.
\newblock In \emph{Proceedings of the 62nd Annual Meeting of the Association for Computational Linguistics (Volume 1: Long Papers)}, pp.\  1658--1677, 2024.

\bibitem[Kang et~al.(2025)Kang, Chen, Han, Inan, Wutschitz, Chen, Sim, and Rajmohan]{kang2025acon}
Minki Kang, Wei-Ning Chen, Dongge Han, Huseyin~A Inan, Lukas Wutschitz, Yanzhi Chen, Robert Sim, and Saravan Rajmohan.
\newblock Acon: Optimizing context compression for long-horizon llm agents.
\newblock \emph{arXiv preprint arXiv:2510.00615}, 2025.

\bibitem[Kim et~al.(2026)Kim, Kim, Kwon, Lee, Yun, and Song]{kim2026kvzip}
Jang-Hyun Kim, Jinuk Kim, Sangwoo Kwon, Jae~W Lee, Sangdoo Yun, and Hyun~Oh Song.
\newblock Kvzip: Query-agnostic kv cache compression with context reconstruction.
\newblock \emph{Advances in Neural Information Processing Systems}, 38:\penalty0 167563--167591, 2026.

\bibitem[Kwon et~al.(2023)Kwon, Li, Zhuang, Sheng, Zheng, Yu, Gonzalez, Zhang, and Stoica]{kwon2023efficient}
Woosuk Kwon, Zhuohan Li, Siyuan Zhuang, Ying Sheng, Lianmin Zheng, Cody~Hao Yu, Joseph Gonzalez, Hao Zhang, and Ion Stoica.
\newblock Efficient memory management for large language model serving with pagedattention.
\newblock In \emph{Proceedings of the 29th symposium on operating systems principles}, pp.\  611--626, 2023.

\bibitem[Laban et~al.(2026)Laban, Hayashi, Zhou, and Neville]{laban2026llms}
Philippe Laban, Hiroaki Hayashi, Yingbo Zhou, and Jennifer Neville.
\newblock Llms get lost in multi-turn conversation.
\newblock In \emph{International Conference on Learning Representations}, volume 2026, pp.\  54738--54778, 2026.

\bibitem[Li et~al.(2026)Li, Lou, and Li]{li2026intentkv}
Junjie Li, Jiong Lou, and Jie Li.
\newblock Intentkv: Cross-turn intent-aware kv cache pruning for agent inference.
\newblock \emph{arXiv preprint arXiv:2606.09916}, 2026.

\bibitem[Li et~al.(2025)Li, Jiang, Wu, Luo, Ahn, Zhang, Abdi, Li, Gao, Yang, et~al.]{li2025scbench}
Yucheng Li, Huiqiang Jiang, Qianhui Wu, Xufang Luo, Surin Ahn, Chengruidong Zhang, Amir Abdi, Dongsheng Li, Jianfeng Gao, Yuqing Yang, et~al.
\newblock Scbench: A kv cache-centric analysis of long-context methods.
\newblock In \emph{International Conference on Learning Representations}, volume 2025, pp.\  66063--66093, 2025.

\bibitem[Li et~al.(2024)Li, Huang, Yang, Venkitesh, Locatelli, Ye, Cai, Lewis, and Chen]{li2024snapkv}
Yuhong Li, Yingbing Huang, Bowen Yang, Bharat Venkitesh, Acyr Locatelli, Hanchen Ye, Tianle Cai, Patrick Lewis, and Deming Chen.
\newblock Snapkv: Llm knows what you are looking for before generation.
\newblock \emph{Advances in Neural Information Processing Systems}, 37:\penalty0 22947--22970, 2024.

\bibitem[Lindenbauer et~al.(2025)Lindenbauer, Slinko, Felder, Bogomolov, and Zharov]{lindenbauer2025complexity}
Tobias Lindenbauer, Igor Slinko, Ludwig Felder, Egor Bogomolov, and Yaroslav Zharov.
\newblock The complexity trap: Simple observation masking is as efficient as llm summarization for agent context management.
\newblock \emph{arXiv preprint arXiv:2508.21433}, 2025.

\bibitem[Liu et~al.(2026)Liu, Ji, An, Jain, Polatkan, Zhu, and Chang]{liu2026practical}
Yujian Liu, Jiabao Ji, Li~An, Rohit Jain, Gungor Polatkan, Siyu Zhu, and Shiyu Chang.
\newblock Practical online kv cache compaction for llm agents: An empirical study.
\newblock \emph{arXiv preprint arXiv:2608.00902}, 2026.

\bibitem[Lu et~al.(2026)Lu, Chen, Fang, Ma, and Wang]{lu2026mix}
Haiquan Lu, Zigeng Chen, Gongfan Fang, Xinyin Ma, and Xinchao Wang.
\newblock Mix-quant: Quantized prefilling, precise decoding for agentic llms.
\newblock \emph{arXiv preprint arXiv:2605.20315}, 2026.

\bibitem[{OpenAI}(2025)]{openai2025codexcli}
{OpenAI}.
\newblock Codex {CLI}: A lightweight coding agent that runs in your terminal.
\newblock \url{https://github.com/openai/codex}, 2025.
\newblock Version X.Y.Z. Accessed: 2026-09-25.

\bibitem[OpenAI et~al.(2025)OpenAI, :, Agarwal, Ahmad, Ai, Altman, Applebaum, Arbus, Arora, Bai, Baker, Bao, Barak, Bennett, Bertao, Brett, Brevdo, Brockman, Bubeck, Chang, Chen, Chen, Cheung, Clark, Cook, Dukhan, Dvorak, Fives, Fomenko, Garipov, Georgiev, Glaese, Gogineni, Goucher, Gross, Guzman, Hallman, Hehir, Heidecke, Helyar, Hu, Huet, Huh, Jain, Johnson, Koch, Kofman, Kundel, Kwon, Kyrylov, Le, Leclerc, Lennon, Lessans, Lezcano-Casado, Li, Li, Lin, Liss, Lily, Liu, Liu, Lu, Lu, Martinovic, McCallum, McGrath, McKinney, McLaughlin, Mei, Mostovoy, Mu, Myles, Neitz, Nichol, Pachocki, Paino, Palmie, Pantuliano, Parascandolo, Park, Pathak, Paz, Peran, Pimenov, Pokrass, Proehl, Qiu, Raila, Raso, Ren, Richardson, Robinson, Rotsted, Salman, Sanjeev, Schwarzer, Sculley, Sikchi, Simon, Singhal, Song, Stuckey, Sun, Tillet, Toizer, Tsimpourlas, Vyas, Wallace, Wang, Wang, Watkins, Weil, Wendling, Whinnery, Whitney, Wong, Yang, Yang, Yasunaga, Ying, Zaremba, Zhan, Zhang, Zhang, Zhang, and
  Zhao]{openai2025gptoss120bgptoss20bmodel}
OpenAI, :, Sandhini Agarwal, Lama Ahmad, Jason Ai, Sam Altman, Andy Applebaum, Edwin Arbus, Rahul~K. Arora, Yu~Bai, Bowen Baker, Haiming Bao, Boaz Barak, Ally Bennett, Tyler Bertao, Nivedita Brett, Eugene Brevdo, Greg Brockman, Sebastien Bubeck, Che Chang, Kai Chen, Mark Chen, Enoch Cheung, Aidan Clark, Dan Cook, Marat Dukhan, Casey Dvorak, Kevin Fives, Vlad Fomenko, Timur Garipov, Kristian Georgiev, Mia Glaese, Tarun Gogineni, Adam Goucher, Lukas Gross, Katia~Gil Guzman, John Hallman, Jackie Hehir, Johannes Heidecke, Alec Helyar, Haitang Hu, Romain Huet, Jacob Huh, Saachi Jain, Zach Johnson, Chris Koch, Irina Kofman, Dominik Kundel, Jason Kwon, Volodymyr Kyrylov, Elaine~Ya Le, Guillaume Leclerc, James~Park Lennon, Scott Lessans, Mario Lezcano-Casado, Yuanzhi Li, Zhuohan Li, Ji~Lin, Jordan Liss, Lily, Liu, Jiancheng Liu, Kevin Lu, Chris Lu, Zoran Martinovic, Lindsay McCallum, Josh McGrath, Scott McKinney, Aidan McLaughlin, Song Mei, Steve Mostovoy, Tong Mu, Gideon Myles, Alexander Neitz, Alex Nichol, Jakub
  Pachocki, Alex Paino, Dana Palmie, Ashley Pantuliano, Giambattista Parascandolo, Jongsoo Park, Leher Pathak, Carolina Paz, Ludovic Peran, Dmitry Pimenov, Michelle Pokrass, Elizabeth Proehl, Huida Qiu, Gaby Raila, Filippo Raso, Hongyu Ren, Kimmy Richardson, David Robinson, Bob Rotsted, Hadi Salman, Suvansh Sanjeev, Max Schwarzer, D.~Sculley, Harshit Sikchi, Kendal Simon, Karan Singhal, Yang Song, Dane Stuckey, Zhiqing Sun, Philippe Tillet, Sam Toizer, Foivos Tsimpourlas, Nikhil Vyas, Eric Wallace, Xin Wang, Miles Wang, Olivia Watkins, Kevin Weil, Amy Wendling, Kevin Whinnery, Cedric Whitney, Hannah Wong, Lin Yang, Yu~Yang, Michihiro Yasunaga, Kristen Ying, Wojciech Zaremba, Wenting Zhan, Cyril Zhang, Brian Zhang, Eddie Zhang, and Shengjia Zhao.
\newblock gpt-oss-120b \& gpt-oss-20b model card, 2025.
\newblock URL \url{https://arxiv.org/abs/2508.10925}.

\bibitem[Pan et~al.(2024)Pan, Wu, Jiang, Xia, Luo, Zhang, Lin, R{\"u}hle, Yang, Lin, et~al.]{pan2024llmlingua}
Zhuoshi Pan, Qianhui Wu, Huiqiang Jiang, Menglin Xia, Xufang Luo, Jue Zhang, Qingwei Lin, Victor R{\"u}hle, Yuqing Yang, Chin-Yew Lin, et~al.
\newblock Llmlingua-2: Data distillation for efficient and faithful task-agnostic prompt compression.
\newblock In \emph{Findings of the Association for Computational Linguistics: ACL 2024}, pp.\  963--981, 2024.

\bibitem[Sun et~al.(2025)Sun, Lu, Ling, Liu, Yao, Yang, and Chen]{sun2025scaling}
Weiwei Sun, Miao Lu, Zhan Ling, Kang Liu, Xuesong Yao, Yiming Yang, and Jiecao Chen.
\newblock Scaling long-horizon llm agent via context-folding.
\newblock \emph{arXiv preprint arXiv:2510.11967}, 2025.

\bibitem[Tang et~al.(2024)Tang, Zhao, Zhu, Xiao, Kasikci, and Han]{tang2024quest}
Jiaming Tang, Yilong Zhao, Kan Zhu, Guangxuan Xiao, Baris Kasikci, and Song Han.
\newblock Quest: Query-aware sparsity for efficient long-context llm inference.
\newblock \emph{arXiv preprint arXiv:2406.10774}, 2024.

\bibitem[Wu et~al.(2026{\natexlab{a}})Wu, Zhang, Ghosh, Basu, Deoras, Huan, and Gupta]{wu2026contextweaver}
Yating Wu, Yuhao Zhang, Sayan Ghosh, Sourya Basu, Anoop Deoras, Jun Huan, and Gaurav Gupta.
\newblock Contextweaver: Selective and dependency-structured memory construction for llm agents.
\newblock \emph{arXiv preprint arXiv:2604.23069}, 2026{\natexlab{a}}.

\bibitem[Wu et~al.(2026{\natexlab{b}})Wu, Zhao, Li, Lee, Zhu, Wu, Yu, Li, Zhang, Fan, et~al.]{wu2026swe}
Yifan Wu, Zhuokai Zhao, Songlin Li, Ho~Hin Lee, Jiacheng Zhu, Shirley Wu, Tianhe Yu, Serena Li, Lizhu Zhang, Xiangjun Fan, et~al.
\newblock Swe-together: Evaluating coding agents in interactive user sessions.
\newblock \emph{arXiv preprint arXiv:2606.29957}, 2026{\natexlab{b}}.

\bibitem[Xiao et~al.(2024)Xiao, Zhang, Han, Xiao, Lin, Zhang, Liu, and Sun]{xiao2024infllm}
Chaojun Xiao, Pengle Zhang, Xu~Han, Guangxuan Xiao, Yankai Lin, Zhengyan Zhang, Zhiyuan Liu, and Maosong Sun.
\newblock Infllm: Training-free long-context extrapolation for llms with an efficient context memory.
\newblock \emph{Advances in neural information processing systems}, 37:\penalty0 119638--119661, 2024.

\bibitem[Yang et~al.(2025)Yang, Li, Yang, Zhang, Hui, Zheng, Yu, Gao, Huang, Lv, et~al.]{yang2025qwen3}
An~Yang, Anfeng Li, Baosong Yang, Beichen Zhang, Binyuan Hui, Bo~Zheng, Bowen Yu, Chang Gao, Chengen Huang, Chenxu Lv, et~al.
\newblock Qwen3 technical report.
\newblock \emph{arXiv preprint arXiv:2505.09388}, 2025.

\bibitem[Zeng et~al.(2026)Zeng, Li, Xie, Ye, and Zhang]{zeng2026attncompress}
Zhengran Zeng, Yixin Li, Rui Xie, Wei Ye, and Shikun Zhang.
\newblock Attncompress: Dynamic attention-guided trajectory compression for software engineering agents.
\newblock \emph{arXiv preprint arXiv:2609.08318}, 2026.

\bibitem[Zhang et~al.(2023)Zhang, Sheng, Zhou, Chen, Zheng, Cai, Song, Tian, R{\'e}, Barrett, et~al.]{zhang2023h2o}
Zhenyu Zhang, Ying Sheng, Tianyi Zhou, Tianlong Chen, Lianmin Zheng, Ruisi Cai, Zhao Song, Yuandong Tian, Christopher R{\'e}, Clark Barrett, et~al.
\newblock H2o: Heavy-hitter oracle for efficient generative inference of large language models.
\newblock \emph{Advances in neural information processing systems}, 36:\penalty0 34661--34710, 2023.

\bibitem[Zheng et~al.(2024)Zheng, Yin, Xie, Sun, Huang, Yu, Cao, Kozyrakis, Stoica, Gonzalez, et~al.]{zheng2024sglang}
Lianmin Zheng, Liangsheng Yin, Zhiqiang Xie, Chuyue Sun, Jeff Huang, Cody~H Yu, Shiyi Cao, Christos Kozyrakis, Ion Stoica, Joseph~E Gonzalez, et~al.
\newblock Sglang: Efficient execution of structured language model programs.
\newblock \emph{Advances in neural information processing systems}, 37:\penalty0 62557--62583, 2024.

\bibitem[Zhou et~al.(2026)Zhou, Qu, Wu, Kim, Prakash, Rus, Low, and Liang]{zhou2026mem1}
Zijian Zhou, Ao~Qu, Zhaoxuan Wu, Sunghwan Kim, Alok Prakash, Daniela Rus, Bryan Kian~Hsiang Low, and Paul Liang.
\newblock Mem1: Learning to synergize memory and reasoning for efficient long-horizon agents.
\newblock In \emph{International Conference on Learning Representations}, volume 2026, pp.\  58413--58438, 2026.

\bibitem[Zhu et~al.(2026)Zhu, Jacob, Ma, Pan, Wang, Krishnamurthy, and Kasikci]{zhu2026tracelab}
Kan Zhu, Mathew Jacob, Chenxi Ma, Yi~Pan, Stephanie Wang, Arvind Krishnamurthy, and Baris Kasikci.
\newblock Tracelab: Characterizing coding agent workloads for llm serving.
\newblock \emph{arXiv preprint arXiv:2606.30560}, 2026.

\bibitem[Zweiger et~al.(2026)Zweiger, Fu, Guo, and Kim]{zweiger2026fast}
Adam Zweiger, Xinghong Fu, Han Guo, and Yoon Kim.
\newblock Fast kv compaction via attention matching.
\newblock \emph{arXiv preprint arXiv:2602.16284}, 2026.

\end{thebibliography}
